\documentclass{article}
\usepackage[preprint]{neurips_2025}

\usepackage[utf8]{inputenc} % allow utf-8 input
\usepackage[T1]{fontenc}    % use 8-bit T1 fonts
\usepackage{hyperref}       % hyperlinks
\usepackage{url}            % simple URL typesetting
\usepackage{booktabs}       % professional-quality tables
\usepackage{amsfonts}       % blackboard math symbols
\usepackage{nicefrac}       % compact symbols for 1/2, etc.
\usepackage{microtype}      % microtypography
\usepackage{xcolor}         % colors

\usepackage{tabularx}
\usepackage{multirow}
\usepackage{makecell}

\title{On the Detectability of LLM-Generated Text: \\ What Exactly Is LLM-Generated Text?}

\author{%
  Mingmeng Geng \qquad Thierry Poibeau \\
  École Normale Supérieure (ENS) - Université Paris Sciences et Lettres (PSL) \\
  Laboratoire Lattice (CNRS, ENS-PSL, Université Sorbonne Nouvelle)\\
  \texttt{mingmeng.geng@ens.psl.eu} \\
}

\begin{document}

\maketitle
\begin{abstract}
With the widespread use of large language models (LLMs), many researchers have turned their attention to detecting text generated by them. However, there is no consistent or precise definition of their target, namely ``LLM-generated text''. Differences in usage scenarios and the diversity of LLMs further increase the difficulty of detection. What is commonly regarded as the detecting target usually represents only a subset of the text that LLMs can potentially produce. Human edits to LLM outputs, together with the subtle influences that LLMs exert on their users, are blurring the line between LLM-generated and human-written text. Existing benchmarks and evaluation approaches do not adequately address the various conditions in real-world detector applications. Hence, the numerical results of detectors are often misunderstood, and their significance is diminishing. Therefore, detectors remain useful under specific conditions, but their results should be interpreted only as references rather than decisive indicators.
\end{abstract}

\section{Introduction}

The rapid development of large language models (LLMs) has led to a rise of LLM-generated text observed across various fields, such as academia~\citep{liang2024monitoring,geng2024chatgpt} and Wikipedia~\citep{brooks2024rise,huang2025wikipedia}. The use of AI may also be associated with negative judgments (e.g., low-quality works and lack of creativity), which has become a social phenomenon~\citep{sarkar2025ai}. Thus, the detection of LLM-generated text is not merely a technical matter, but also a social issue. 

A lot of detectors have been proposed and studied~\citep{yang2023survey,wu2025survey}, while various issues and challenges have also been examined~\citep{tang2024science,wu2025survey,fraser2024detecting,abdali2024decoding}. The reliability of the detectors has also been widely discussed, for example, their robustness against attacks and their use in different situations~\citep{sadasivan2023can,ghosal2023towards,chakraborty2024position}. Similar to \citet{chakraborty2024position}'s paper, we explore the following question:
\begin{center}
    \textit{Is it possible to detect the LLM-generated text in practice?}
\end{center}
Researchers often say that the text generated by LLMs is different from that written by humans~\citep{pu2023chatgpt,munoz2024contrasting,reinhart2025llms}. The indistinguishability between LLM-generated and human-written text is one of big challenges for LLMs~\citep{kaddour2023challenges}. However, a basic problem has not been adequately addressed in current literature: 
\begin{center}
    \textit{What exactly is ``LLM-generated text''?}
\end{center}
 A simpler and more direct example is that the output of LLMs will vary when given different prompts. Just as different people can write in various styles~\citep{zhu2021idiosyncratic}, different LLMs generate text in different styles~\citep{rosenfeld2024whose,sun2025idiosyncrasies}. 

The broad and vague definitions of LLM-generated text have resulted in inconsistent benchmarks and evaluations, which, in turn, has created new ethical issues and challenges in practical use. If we take into account the diversity of LLMs and human-in-the-loop, the situation becomes even more intricate, and few detectors have tried to distinguish them~\citep{cheng2025beyond}. It has shown that humans possess the capacity to identify LLM-generated text with reasonable accuracy~\citep{russell2025people}, and the coevolution between humans and LLMs may have occurred~\citep{geng2025human}. Empirical studies have consistently demonstrated that different LLMs exhibit distinct stylistic patterns fingerprints, which could even be used for classifiction~\citep{mcgovern2024your,sun2025idiosyncrasies,bitton2025detecting}.  The continuous evolution and adjustment of watermarking and attacks are also related to its definition, adding further complexity to the issue of detectability. LLMs may be fully capable of generating text that is nearly indistinguishable from human-written content in the future. 

Hence, in many practical cases, we think that it is not possible to accurately detect LLM-generated text. We will discuss the detectability of LLM-generated text in detail in the following sections.

\section{Definition and Usage}
As we mentioned earlier, what exactly is the detection target: ``LLM-generated text''? Researchers have referred to it using expressions like ``machine-generated text'' or ``AI-generated'', as listed in Table~\ref{tab_definition}. For simplicity, we use ``LLM-generated text'' to represent the subject of study in this paper, as it is more precise than the other expressions. 

\begin{table}[htbp]
  \centering
  \begin{tabularx}{\textwidth}{lX}
  \toprule
    \textbf{Paper}           & \textbf{Definition} \\
    \midrule
    \citet{crothers2023machine} & \textit{``\underline{Machine-generated text}'' is natural language text that is produced, modified, or extended by a machine.}\\
    \citet{kumarage2024survey} & \textit{In this survey, we define \underline{AI-generated text} as output produced by a natural language generation pipeline employing a neural probabilistic language model}. \\
    \citet{wu2025survey} & \textit{\underline{LLM-generated Text} is defined as cohesive, grammatically sound, and pertinent content generated by LLMs}. \\ 
  \bottomrule
  \end{tabularx}
  \vspace{5pt}
  \caption{Definition of LLM-generated text in different papers}
  \label{tab_definition}
\end{table}

 Apart from slight differences in terminology, the definition of ``LLM-generated text'' in most papers is quite broad, meaning the text can be produced in many ways using LLMs, like paraphrasing, translation, or generating long text from simple prompts. Theoretically, the difference between ``LLM-generated text'' and ``human-written text'' lies in how they are produced, as implied by the names. But in practice, we can only evaluate them based on their final output i.e., the text, where a considerable overlap exists between them.

We think that \textbf{what is commonly regarded as the detecting target usually represents only a subset of the text that LLMs can potentially produce}, and it’s likely the kind that corresponds to the most common and direct prompts. For instance, many detectors are trained on text generated by LLMs, which cannot represent all possibilities. Consequently, their detection capabilities are constrained. While different parameters can be set for various types of cases~\citep{hans2024spotting}, such configurations can hardly cover all possible scenarios.

In addition to the examples given earlier, the traces of LLM-generated text have now been found in various fields, such as student essays' answers~\citep{leppanen2025large}, words used in speaking~\citep{yakura2024empirical,geng2024impact,galpin2025exploring} and numerous online texts~\citep{sun2024we,liang2025widespread,la2025machines}. We can even say that, LLM-generated text has now integrated into all areas of life, mixing with human writing.

While limitations of these detection methods have caused concern among researchers ~\citep{sadasivan2023can,liang2023gpt,doughman2024exploring,nicks2023language,saha2025almost}, they could be applied in diverse contexts. Such detection tools are often promoted for their potential utility in identifying instances of plagiarism~\citep{pudasaini2024survey}, academic dishonesty (e.g., cheating during examinations)~\citep{wang2025continually}, the automatic generation of unethical peer reviews~\citep{kumar2025mixrevdetect}, and other forms of content manipulation.

Therefore, evaluating and comparing the effectiveness of different detectors presents a significant challenge.

\section{Background}

Probably most people became aware of LLMs after the release of ChatGPT, but the research on detecting text generated by language models had started before that. For example, \citet{gehrmann2019gltr} proposed the GLTR (Giant Language model Test Room) tool to detect whether text was generated by models, with experiments involving GPT-2~\citep{radford2019language} and BERT~\citep{devlin2019bert}. \citet{zellers2019defending} developed the Grover model to detect AI-generated fake news. Even GPT-3~\citep{brown2020language} continued to face skepticism regarding its text-generation capabilities~\citep{bender2021dangers}, making the detectors' performance unsurprising. Another pioneering work by \citet{ippolito2019automatic} demonstrated that humans have already encountered some difficulties in identifying texts generated by GPT-2. Later, \citet{clark2021all} found that untrained people at the time were not very good at recognizing text produced by GPT-3, and \citet{wahle2022large} noticed the similar situation for machine-paraphrased plagiarism.

The rapid development and spread of LLMs has drawn significant attention from researchers to the detection of LLM-generated text, and diverse methods have been proposed~\citep{wu2025survey}: DetectGPT~\citep{mitchell2023detectgpt}, Fast-DetectGPT~\citep{bao2023fast}, DetectLLM~\citep{su2023detectllm}, LLMDet~\citep{wu2023llmdet}, DeID-GPT~\citep{liu2023deid} and some others in 2023~\citep{dugan2023real,zhu2023beat}; Binoculars~\citep{hans2024spotting}, TOCSIN~\citep{ma2024zero}, Dpic~\citep{yu2024dpic}, Text Fluoroscopy~\citep{yu2024text} in 2024, some more detecting methods in 2025~\citep{zhou2025adadetectgpt,zeng2025human}. The examples listed above are illustrative, and the actual number of detectors is much larger.  

These detection methods can be classified into many categories according to different criteria. For instance, \citet{abdali2024decoding} classifies them as supervised methods, zero-shot methods, retrieval-based methods, watermarking methods, discriminating features. \citet{wu2025survey} mainly examines them through the lens of watermarking techniques, statistics-based detectors, neural-based detectors, and human-assisted methods. Specialized detectors have been developed, for instance, targeting tweets~\citep{gambini2022pushing,kumarage2023stylometric}, journalistic news articles~\citep{bhattacharjee2023conda} and Wikipedia~\citep{quaremba2025wetbench}. Additionally, the detection of LLM-generated text is not limited to English~\citep{wang2025genai}. Detectors that are used in other languages have also been brought up, such as French~\citep{antoun2023towards}, Japanese~\citep{zaitsu2023distinguishing}, Chinese~\citep{wang2024llm}.

There are other ways to categorize the detectors. For example, most studies only think about binary classification, and detectors with multi-category cases have not been comprehensively explored. In spite of the wide range of methods utilized by these detectors, the absence of universal benchmarks and different application scenarios limits a meaningful comparison~\citep{dugan2024raid}. Although these detection techniques performed well earlier on certain datasets, the ongoing progress of LLMs also makes detection harder~\citep{wu2025survey}.

Researchers have not yet reached full agreement on the detectability of LLM-generated text. For instance, \citet{chakraborty2024position} claim in their position paper: ``\textit{Despite ongoing debate about the feasibility of such differentiation, we present evidence supporting its consistent achievability, except when human and machine text distributions are indistinguishable across their entire support. Drawing from information theory, we argue that as machine-generated text approximates human-like quality, the sample size needed for detection increases.}'' While \citet{zhang2024detection} argue that "\textit{We believe that the issue of AI-generated text detection remains an unresolved challenge. As LLMs become increasingly powerful and humans become more proficient in using them, it is even less likely to detect AI text in the future.}" \citet{nicks2023language} ``\textit{advise against continued reliance on LLM-generated text detectors}'', and \citet{ganie2025uncertainty} try to show that ``\textit{perfect AI detection is mathematically impossible}''.

The key disagreement among researchers may not be technical in nature, but rather stems from differing perspectives on human intervention and usage scenarios. The assumptions made by different researchers on this issue are not the same. Take watermarking studies as an example, if people edit the generated text (which is simple to do), the watermark’s reliability may be greatly weakened~\citep{dathathri2024scalable}. 

\section{Evaluation and Benchmark}

Now, we will focus on a relatively simple scenario, i.e., we consider that \textbf{``LLM-generated text'' refers to the content produced by LLMs through normal prompts}.

As outlined earlier, there have always been many doubts about the effectiveness of these detectors~\citep{sadasivan2023can,weber2023testing}. The sensitivity of detectors and the problem of false positives have been widely criticized and examined in various papers~\citep{weber2023testing,walters2023effectiveness,kar2024sensitive,tufts2024practical}. By way of illustration, a detection accuracy of 85\% necessarily implies a 15\% error rate, which may include both false positives and false negatives. \citet{rashidi2023chatgpt} find that the AI text detector erroneously identified up to 8\% of the known real abstracts as AI-generated text, and \citet{giray2024problem} states that false positives disproportionately affect non-native English speakers and scholars with distinctive writing styles. 

Studies indicate that the detectability of texts depends on the LLM used for text generation~\citep{antoun2023text}. For instance, \citet{wu2024detectrl} point out that the Binoculars~\citep{hans2024spotting} only achieved a 55.15\% AUROC in detecting texts generated by Claude, while for texts generated by several other models, it reached at least 88\%. Given the diversity of usage scenarios and the variety of LLMs, such results are to be expected and no existing LLM-detection system is infallible.

Besides, most studies focus on the binary classification problem of determining whether a given text is generated by LLMs. While some detection methods could achieve good results on given datasets, the scenario becomes more much complicated in real-world settings. To exemplify, people could edit LLM-generated text or mix it with human written text, which has also attracted considerable attention~\citep{zhang2024llm,abassy2024llm,kumar2025mixrevdetect,kadiyala2025robust}. Human contribution also affects detection results~\citep{gehring2025assessing}. Only a small number of researchers have tried to identify specific roles of LLM in content generation~\citep{cheng2025beyond}, and no universally accepted approaches have been established. 

Another challenge in establishing benchmarks for detecting LLM-generated text is that LLMs are continuously evolving, and their characteristics do not remain the same. To cite an example, \citep{liyanage2022benchmark} create their benchmark with GPT-2, which should be quite differently from the current advanced LLMs. Although LLMs were compared to stochastic parrot~\citep{bender2021dangers} a couple of years ago, their capabilities are gradually being recognized~\citep{srivastava2022beyond}, and now their competencies have reached or even surpassed those of human experts in various fields. This leads us to speculate that future LLM-generated texts could surpass current versions in human resemblance. The persona effect could be a good example~\citep{hu2024quantifying,li2025writes,gao2025personalization}.

Some researchers have shown that detectors may more easily flag text from older and smaller models~\citet{elkhatat2023evaluating,saha2025almost}. The performance of detection systems varies with the application scenario; for instance, they tend to be less effective when dealing with short texts~\citep{gameiro2024llm,shi2024ten}. More LLMs were employed in subsequent benchmark construction~\citep{wang2024m4gt,he2024mgtbench, cornelius2024bust}, but the number of prompts and scenarios used was limited. Some recent benchmarks~\citep{tao2024cudrt,wu2024detectrl} have incorporated a broader range of scenarios, and their impact and effectiveness remain to be seen. There is a similar issue with datasets. For example, \citet{gritsai2024ai} found that all analyzed datasets fail in one or another of their methods and do not allow to reliably estimate AI detectors. 

Therefore, their adoption has not yet become as widespread as other well-known LLM benchmarks, such as GLUE~\citep{wang2018glue} and MMLU~\citep{hendrycks2020measuring}. Although these benchmarks have also faced some criticism~\citep{hadi2023survey}, there is still no highly universal benchmark to detect LLM-generated texts. Researchers have also started developing shared benchmarks to tackle this problem~\citep{dugan2024raid}. Nevertheless, due to the diversity of usage scenarios and the continuous updates of LLMs, a gold-standard benchmark is hard to realize, may even remain permanently absent.

\section{Attack, Watermarking, and Evolution}

The brittleness of these tools is demonstrated by many realistic scenarios, including various attacks. An early study has shown that while humans can reliably detect poetry produced by GPT-2, but they struggle to accurately recognize it after human selection~\citep{kobis2021artificial}. If post-generation modifications are taken into account, the detection process should become more challenging. For example, although some recent systems claim to maintain performance when the LLM-generated text is lightly modified, empirical evidence shows that detection accuracy tends to decline as the extent of human revision increases~\citep{lau2024understanding}. 

Studies on how to attack these detectors are also being conducted in different views, such as paraphrasing~\citep{sadasivan2023can,krishna2023paraphrasing}, adversarial prompting~\citep{tufts2024practical,cheng2025adversarial}, and other forms of authorship obfuscation~\citep{macko2024authorship}. \citet{tufts2024practical} think that adversarial attacks can easily bypass these detectors, and balancing high sensitivity with a reasonable true positive rate remains challenging. \citet{dubois2025sampling} found that even minor adjustments to decoding parameters (such as temperature and top-p) can severely impair detector accuracy. The text produced by the fine-tuned model could also be more challenging to detect~\citep{dawkins2025detection}. Some subtle differences could also affect the detectors’ results, such as a single space~\citep{cai2023evade} or some minor perturbations~\citep{zhou2024humanizing}. 

To address the detection problem at its source, researchers have already begun experimenting with watermarking methods, which has shown good effectiveness in simulation~\citep{kirchenbauer2023watermark} and may also be an ethical necessity~\citep{grinbaum2022ethical}. Diverse watermarking techniques have been proposed, in theory~\citep{christ2024undetectable} and in practice~\citep{dathathri2024scalable}. In addition, some researchers have pointed out the potential risk of watermark stealing~\citep{jovanovic2024watermark}, and studies on adversarial attacks against watermarking are ongoing~\citep{xie2025detecting}.

While there are specific solutions for certain attacks~\citep{huang2024ai,fang2025your}, in practice, there are numerous methods of attack. Consequently, the tools’ applicability to real-world use cases remains limited. This limitation exacerbates the concerns raised in the first point, as reliance on imperfect systems in nuanced or ambiguous situations increases the likelihood of erroneous judgments. 

And some of these difficulties are simply unavoidable, not merely temporary technical challenges. As noted by several researchers, humans can learn to detect AI-generated texts~\citep{milivcka2025humans} and can become an accurate and robust detector of LLM-generated text~\citep{russell2025people}. When LLMs become more widely used and people may learn from their outputs, the difference between them may get further smaller. We can find more cases of coevolution~\citep{pedreschi2024human,geng2025human} between human and machines. Therefore, the gap between LLM-generated and human-written text is expected to narrow further, but we think that they will not become entirely identical.

Consequently, in the future, detecting LLM-generated text may become less important, particularly in terms of numerical interpretation. While the hybrid human-in-the-loop model is a possible option for detectors, it should be regarded as one way to mitigate the problem, not a perfect solution. 

\section{Ethical Considerations}
The social impact of of LLMs has already been considered~\citep{solaiman2019release}. People began discussing their positive impact shortly after its emergence~\citep{kasneci2023chatgpt}. LLMs can play a valuable role in numerous contexts, such as enhancing productivity~\citep{noy2023experimental}. LLMs can help bridge linguistic divides by enabling non-native speakers to produce coherent and idiomatic texts in English or other target languages, thereby supporting greater inclusivity in academic and professional communication. Non-native English speakers have to put in more effort as scientists, and there has been discrimination in the past~\citep{amano2023manifold,lepp2025you}. Automatic editing methods have shown promise in improving alignment between LLM-generated and human-written text~\citep{chakrabarty2024can}. LLMs may have the potential to serve as effective tutors in English Education~\citep{ye2025position}. 

However, text generated with LLMs is frequently the subject of criticism~\citep{sarkar2025ai}, particularly due to concerns about academic dishonesty, plagiarism, and fraud, which has led to the development of various detection tools~\citep{wahle2022large,wu2025survey,iofinova2025position}. As we have discussed, the definition of LLM-generated text is broad and lacks consistency. In some cases, the LLM acts as a language aid, and the substantive intellectual work remains the responsibility of human authors. It is therefore natural to ask the following question:
\begin{center}
    \textit{Should we use these detectors?}
\end{center}
In addition to the technical challenges we have previously discussed, the use of detectors requires consideration of possible ethical risks. A non-native speaker might use an LLM to translate, rephrase, or refine a manuscript that is otherwise the product of original research. But experiments show that certain detectors may exhibit bias against non-native English writers~\citep{liang2023gpt} or against certain demographic groups~\citep{kadoma2025generative}. The analyses from \citet{li2025writes} reveal that all the detectors they tested are highly sensitive to CEFR level and language environment.  Signs of language discrimination and resistance have also been noted in recently~\citep{lepp2025you}. 

As such, current detectors are far from perfect and may never achieve perfection in the future either. Accusing someone of misconduct based on an imperfect tool can lead to unjust outcomes, reputational damage, and institutional distrust. Therefore, even detectors with relatively high accuracy present significant risks when used for evaluative or disciplinary purposes. With LLMs being so widely used in academia~\citep{eger2025transforming,russell2025people}, detecting AI-generated text must be handled with extreme care.

In addition to the detection and policing of LLM-generated text, it may also be useful to advocate for transparency regarding their use~\citep{hosseini2023ethics} and promote AI literacy~\citep{cardon2023challenges}. In academic publishing, for example, it is increasingly common to disclose how LLMs have assisted in drafting, editing, or rephrasing portions of a manuscript. Clear guidelines and disclosures can thus help normalize the ethical integration of LLMs into scholarly workflows without undermining academic integrity.

If LLMs are applied properly and people assess detection tools reasonably, their positive influence can be greatly amplified. Those familiar with LLMs and detectors are aware of the potential issues, but the public tends to be easily drawn to these numbers and the brief conclusion. The mechanism behind the language style of LLMs is currently unclear~\citep{juzek2024does}. The lack of detector interpretability represents another concern~\citep{ji2024detecting}, severely limiting the ability to provide transparent explanations to the public.

\section{Case Study}

We have previously reviewed numerous problems with the detection results of LLM-generated text, and new studies keep confirming these issues~\citep{pudasaini2025benchmarking,pedrotti2025stress,zhang2025character}. Therefore, we will illustrate this with simple examples using different LLMs (DeepSeek-V3.2~\citep{liu2024deepseek}, DeepSeek-R1~\citep{guo2025deepseek}, GPT-3.5, GPT-4o-mini, GPT-4o~\citep{hurst2024gpt}). The prompts employed are as follows:
\begin{itemize}
    \item \textbf{(P1)} Polish the following passage (provide only the result): \dots
    \item \textbf{(P2)} Rewrite the following passage (provide only the result): \dots
    \item \textbf{(P3)} Please polish the following passage (provide only the result): \dots
    \item \textbf{(P4)} Please rewrite the following passage (provide only the result): \dots
\end{itemize}

The original text (the first paragraph of Alan Turing's famous paper \textit{Computing Machinery and Intelligence}~\citep{turing2007computing}) and a processed example are shown below. More results are provided in the Appendix \ref{simulation_results}. 

\paragraph{Original Text} \textit{I propose to consider the question, "Can machines think?" This should begin with definitions of the meaning of the terms "machine" and "think." The definitions might be framed so as to reflect so far as possible the normal use of the words, but this attitude is dangerous, If the meaning of the words "machine" and "think" are to be found by examining how they are commonly used it is difficult to escape the conclusion that the meaning and the answer to the question, "Can machines think?" is to be sought in a statistical survey such as a Gallup poll. But this is absurd. Instead of attempting such a definition I shall replace the question by another, which is closely related to it and is expressed in relatively unambiguous words.}

\paragraph{GPT-4o \& (P1):} \textit{I propose to consider the question, "Can machines think?" This should begin with defining the terms "machine" and "think." While these definitions might aim to reflect the common usage of the words, this approach is risky. If we determine the meanings of "machine" and "think" based on their everyday use, we might conclude that the answer to "Can machines think?" should be derived from a statistical survey, like a Gallup poll. However, this is absurd. Instead of attempting such a definition, I will replace the question with another, closely related one, expressed in relatively unambiguous terms.}

Table~\ref{detection_results} presents the results of detecting these texts using Fast-DetectGPT~\citep{bao2023fast}. Even though these texts are all generated by LLMs, their detection outcomes vary widely. It is not difficult to see that in many cases, compared with the original text, the text processed by LLMs is considered by the detector to appear less machine-generated.

\begin{table}[htbp]
\centering
\begin{tabular}{cccccc}
\toprule
Prompts & (Original text) & P1 & P2 & P3  & P4 \\ 
\midrule
 GPT-3.5 & \multirow{5}{*}{33\% / 17\%}  & \textbf{27\%}  / \textbf{14\%} & 40\%  / 42\%   &  79\%  / 89\% &  \textbf{22\%}  / 23\% \\
 GPT-4o-mini &  & 62\%  / 47\% & \textbf{27\%} / 14\% & 44\%  / 34\% & \textbf{27\%} / \textbf{14\%}\\
 GPT-4o &  & \textbf{26\%}  / \textbf{11\%} & \textbf{28\%}  / 18\% & \textbf{24\%}  / \textbf{12\%} & \textbf{19\%}  / \textbf{11\%}  \\
 DeepSeek-V3.2 &  & 64\% / 54\%  & 50\% / 79\%  & \textbf{32\%} / 18\%   & 60\% / 56\%  \\
 DeepSeek-R1 &  & 43\% / 23\%  & 72\% / 81\%  & 33\% / \textbf{13\%}  & 24\% / \textbf{13\%}  \\
\bottomrule
\end{tabular}
\vspace{5pt}
\caption{Detection results using Fast-DetectGPT. The two values in each cell correspond to the predictions of the machine-generated results when the Sampling/scoring model is gpt-neo-2.7b and falcon-7b, respectively.}
\label{detection_results}
\end{table}
These are merely a few basic examples of the issues and limitations faced by LLM-generated text detectors. Obviously, the same LLM can produce different texts in response to different prompts for the same task, as we have shown before. Although these detectors may still be applicable in certain scenarios, their use requires greater caution.

\section{Conclusions}
We must highlight that a major source of difficulty is that LLM-generated text lacks a unified and clear definition. Moreover, due to the various influences of LLMs on humans, human-written text is likely to converge toward the LLM-generated text. The detection of LLM-generated text can be achieved under multiple assumptions, but in reality, these assumptions are often difficult to satisfy. The abuse and misuse of these detectors can create many risks.

The detectors usually struggle to assess the proportion, function, or ethical significance of LLM contributions in a given text. As a result, the development and deployment of LLM-generated text detectors raise serious concerns, not only due to technical limitations but also because they risk enforcing overly simplistic and potentially unjust frameworks for evaluating authorship and intellectual responsibility.

The numerical effectiveness of LLM-generated text detectors is declining. On the one hand, human may be influenced by LLMs and may create text resembling LLM-generated text. One the other hand, people may also adapt their language to bypass LLM detection tools~\citep{geng2025human}. Detection efforts should target substantive content (e.g., fact-checking) rather than linguistic characteristics~\citep{schuster2020limitations}. 

Therefore, while these detectors can be useful under specific conditions, their results should be used only as a reference with great caution and not as a decisive indicator. For instance, when interpreting the detection results of LLM-generated text, it is necessary to explicitly state the assumptions and prerequisites. 

\section*{Acknowledgements}
This work is supported in part by the ENS-PSL BeYs Chair in Data Science and Cybersecurity. It was also funded in part by the French government under management of Agence Nationale de la Recherche as part of the ``Investissements d'avenir'' program, reference ANR-19-P3IA-0001 (PRAIRIE 3IA Institute).

\bibliographystyle{plainnat}
\bibliography{custom}

\begin{thebibliography}{143}
\providecommand{\natexlab}[1]{#1}
\providecommand{\url}[1]{\texttt{#1}}
\expandafter\ifx\csname urlstyle\endcsname\relax
  \providecommand{\doi}[1]{doi: #1}\else
  \providecommand{\doi}{doi: \begingroup \urlstyle{rm}\Url}\fi

\bibitem[Abassy et~al.(2024)Abassy, Elozeiri, Aziz, Ta, Tomar, Adhikari, Ahmed, Wang, Afzal, Xie, et~al.]{abassy2024llm}
Mervat Abassy, Kareem Elozeiri, Alexander Aziz, Minh~Ngoc Ta, Raj~Vardhan Tomar, Bimarsha Adhikari, Saad El~Dine Ahmed, Yuxia Wang, Osama~Mohammed Afzal, Zhuohan Xie, et~al.
\newblock Llm-detectaive: a tool for fine-grained machine-generated text detection.
\newblock \emph{arXiv preprint arXiv:2408.04284}, 2024.

\bibitem[Abdali et~al.(2024)Abdali, Anarfi, Barberan, and He]{abdali2024decoding}
Sara Abdali, Richard Anarfi, CJ~Barberan, and Jia He.
\newblock Decoding the ai pen: Techniques and challenges in detecting ai-generated text.
\newblock In \emph{Proceedings of the 30th ACM SIGKDD Conference on Knowledge Discovery and Data Mining}, pages 6428--6436, 2024.

\bibitem[Amano et~al.(2023)Amano, Ram{\'\i}rez-Casta{\~n}eda, Berdejo-Espinola, Borokini, Chowdhury, Golivets, Gonz{\'a}lez-Trujillo, Monta{\~n}o-Centellas, Paudel, White, et~al.]{amano2023manifold}
Tatsuya Amano, Valeria Ram{\'\i}rez-Casta{\~n}eda, Violeta Berdejo-Espinola, Israel Borokini, Shawan Chowdhury, Marina Golivets, Juan~David Gonz{\'a}lez-Trujillo, Flavia Monta{\~n}o-Centellas, Kumar Paudel, Rachel~Louise White, et~al.
\newblock The manifold costs of being a non-native english speaker in science.
\newblock \emph{PLoS Biology}, 21\penalty0 (7):\penalty0 e3002184, 2023.

\bibitem[Antoun et~al.(2023{\natexlab{a}})Antoun, Mouilleron, Sagot, and Seddah]{antoun2023towards}
Wissam Antoun, Virginie Mouilleron, Beno{\^\i}t Sagot, and Djam{\'e} Seddah.
\newblock Towards a robust detection of language model generated text: is chatgpt that easy to detect?
\newblock \emph{arXiv preprint arXiv:2306.05871}, 2023{\natexlab{a}}.

\bibitem[Antoun et~al.(2023{\natexlab{b}})Antoun, Sagot, and Seddah]{antoun2023text}
Wissam Antoun, Beno{\^\i}t Sagot, and Djam{\'e} Seddah.
\newblock From text to source: Results in detecting large language model-generated content.
\newblock \emph{arXiv preprint arXiv:2309.13322}, 2023{\natexlab{b}}.

\bibitem[Bao et~al.(2023)Bao, Zhao, Teng, Yang, and Zhang]{bao2023fast}
Guangsheng Bao, Yanbin Zhao, Zhiyang Teng, Linyi Yang, and Yue Zhang.
\newblock Fast-detectgpt: Efficient zero-shot detection of machine-generated text via conditional probability curvature.
\newblock \emph{arXiv preprint arXiv:2310.05130}, 2023.

\bibitem[Bender et~al.(2021)Bender, Gebru, McMillan-Major, and Shmitchell]{bender2021dangers}
Emily~M Bender, Timnit Gebru, Angelina McMillan-Major, and Shmargaret Shmitchell.
\newblock On the dangers of stochastic parrots: Can language models be too big?
\newblock In \emph{Proceedings of the 2021 ACM conference on fairness, accountability, and transparency}, pages 610--623, 2021.

\bibitem[Bhattacharjee et~al.(2023)Bhattacharjee, Kumarage, Moraffah, and Liu]{bhattacharjee2023conda}
Amrita Bhattacharjee, Tharindu Kumarage, Raha Moraffah, and Huan Liu.
\newblock Conda: Contrastive domain adaptation for ai-generated text detection.
\newblock \emph{arXiv preprint arXiv:2309.03992}, 2023.

\bibitem[Bitton et~al.(2025)Bitton, Bitton, and Nisan]{bitton2025detecting}
Yehonatan Bitton, Elad Bitton, and Shai Nisan.
\newblock Detecting stylistic fingerprints of large language models.
\newblock \emph{arXiv preprint arXiv:2503.01659}, 2025.

\bibitem[Brooks et~al.(2024)Brooks, Eggert, and Peskoff]{brooks2024rise}
Creston Brooks, Samuel Eggert, and Denis Peskoff.
\newblock The rise of ai-generated content in wikipedia.
\newblock \emph{arXiv preprint arXiv:2410.08044}, 2024.

\bibitem[Brown et~al.(2020)Brown, Mann, Ryder, Subbiah, Kaplan, Dhariwal, Neelakantan, Shyam, Sastry, Askell, et~al.]{brown2020language}
Tom Brown, Benjamin Mann, Nick Ryder, Melanie Subbiah, Jared~D Kaplan, Prafulla Dhariwal, Arvind Neelakantan, Pranav Shyam, Girish Sastry, Amanda Askell, et~al.
\newblock Language models are few-shot learners.
\newblock \emph{Advances in neural information processing systems}, 33:\penalty0 1877--1901, 2020.

\bibitem[Cai and Cui(2023)]{cai2023evade}
Shuyang Cai and Wanyun Cui.
\newblock Evade chatgpt detectors via a single space.
\newblock \emph{arXiv preprint arXiv:2307.02599}, 2023.

\bibitem[Cardon et~al.(2023)Cardon, Fleischmann, Aritz, Logemann, and Heidewald]{cardon2023challenges}
Peter Cardon, Carolin Fleischmann, Jolanta Aritz, Minna Logemann, and Jeanette Heidewald.
\newblock The challenges and opportunities of ai-assisted writing: Developing ai literacy for the ai age.
\newblock \emph{Business and Professional Communication Quarterly}, 86\penalty0 (3):\penalty0 257--295, 2023.

\bibitem[Chakrabarty et~al.(2024)Chakrabarty, Laban, and Wu]{chakrabarty2024can}
Tuhin Chakrabarty, Philippe Laban, and Chien-Sheng Wu.
\newblock Can ai writing be salvaged? mitigating idiosyncrasies and improving human-ai alignment in the writing process through edits.
\newblock \emph{arXiv preprint arXiv:2409.14509}, 2024.

\bibitem[Chakraborty et~al.(2024)Chakraborty, Bedi, Zhu, An, Manocha, and Huang]{chakraborty2024position}
Souradip Chakraborty, Amrit Bedi, Sicheng Zhu, Bang An, Dinesh Manocha, and Furong Huang.
\newblock Position: On the possibilities of ai-generated text detection.
\newblock In \emph{Forty-first International Conference on Machine Learning}, 2024.

\bibitem[Cheng et~al.(2025{\natexlab{a}})Cheng, Sadasivan, Saberi, Saha, and Feizi]{cheng2025adversarial}
Yize Cheng, Vinu~Sankar Sadasivan, Mehrdad Saberi, Shoumik Saha, and Soheil Feizi.
\newblock Adversarial paraphrasing: A universal attack for humanizing ai-generated text.
\newblock \emph{arXiv preprint arXiv:2506.07001}, 2025{\natexlab{a}}.

\bibitem[Cheng et~al.(2025{\natexlab{b}})Cheng, Zhou, Jiang, Wang, and Li]{cheng2025beyond}
Zihao Cheng, Li~Zhou, Feng Jiang, Benyou Wang, and Haizhou Li.
\newblock Beyond binary: Towards fine-grained llm-generated text detection via role recognition and involvement measurement.
\newblock In \emph{Proceedings of the ACM on Web Conference 2025}, pages 2677--2688, 2025{\natexlab{b}}.

\bibitem[Christ et~al.(2024)Christ, Gunn, and Zamir]{christ2024undetectable}
Miranda Christ, Sam Gunn, and Or~Zamir.
\newblock Undetectable watermarks for language models.
\newblock In \emph{The Thirty Seventh Annual Conference on Learning Theory}, pages 1125--1139. PMLR, 2024.

\bibitem[Clark et~al.(2021)Clark, August, Serrano, Haduong, Gururangan, and Smith]{clark2021all}
Elizabeth Clark, Tal August, Sofia Serrano, Nikita Haduong, Suchin Gururangan, and Noah~A Smith.
\newblock All that's' human'is not gold: Evaluating human evaluation of generated text.
\newblock \emph{arXiv preprint arXiv:2107.00061}, 2021.

\bibitem[Cornelius et~al.(2024)Cornelius, Lithgow-Serrano, Mitrovi{\'c}, Dolamic, and Rinaldi]{cornelius2024bust}
Joseph Cornelius, Oscar Lithgow-Serrano, Sandra Mitrovi{\'c}, Ljiljana Dolamic, and Fabio Rinaldi.
\newblock Bust: Benchmark for the evaluation of detectors of llm-generated text.
\newblock In \emph{Proceedings of the 2024 Conference of the North American Chapter of the Association for Computational Linguistics: Human Language Technologies (Volume 1: Long Papers)}, pages 8029--8057, 2024.

\bibitem[Crothers et~al.(2023)Crothers, Japkowicz, and Viktor]{crothers2023machine}
Evan~N Crothers, Nathalie Japkowicz, and Herna~L Viktor.
\newblock Machine-generated text: A comprehensive survey of threat models and detection methods.
\newblock \emph{IEEE Access}, 11:\penalty0 70977--71002, 2023.

\bibitem[Dathathri et~al.(2024)Dathathri, See, Ghaisas, Huang, McAdam, Welbl, Bachani, Kaskasoli, Stanforth, Matejovicova, et~al.]{dathathri2024scalable}
Sumanth Dathathri, Abigail See, Sumedh Ghaisas, Po-Sen Huang, Rob McAdam, Johannes Welbl, Vandana Bachani, Alex Kaskasoli, Robert Stanforth, Tatiana Matejovicova, et~al.
\newblock Scalable watermarking for identifying large language model outputs.
\newblock \emph{Nature}, 634\penalty0 (8035):\penalty0 818--823, 2024.

\bibitem[Dawkins et~al.(2025)Dawkins, Fraser, and Kiritchenko]{dawkins2025detection}
Hillary Dawkins, Kathleen~C Fraser, and Svetlana Kiritchenko.
\newblock When detection fails: The power of fine-tuned models to generate human-like social media text.
\newblock \emph{arXiv preprint arXiv:2506.09975}, 2025.

\bibitem[Devlin et~al.(2019)Devlin, Chang, Lee, and Toutanova]{devlin2019bert}
Jacob Devlin, Ming-Wei Chang, Kenton Lee, and Kristina Toutanova.
\newblock Bert: Pre-training of deep bidirectional transformers for language understanding.
\newblock In \emph{Proceedings of the 2019 conference of the North American chapter of the association for computational linguistics: human language technologies, volume 1 (long and short papers)}, pages 4171--4186, 2019.

\bibitem[Doughman et~al.(2024)Doughman, Afzal, Toyin, Shehata, Nakov, and Talat]{doughman2024exploring}
Jad Doughman, Osama~Mohammed Afzal, Hawau~Olamide Toyin, Shady Shehata, Preslav Nakov, and Zeerak Talat.
\newblock Exploring the limitations of detecting machine-generated text.
\newblock \emph{arXiv preprint arXiv:2406.11073}, 2024.

\bibitem[Dubois et~al.(2025)Dubois, Yvon, and Piantanida]{dubois2025sampling}
Matthieu Dubois, Fran{\c{c}}ois Yvon, and Pablo Piantanida.
\newblock How sampling affects the detectability of machine-written texts: A comprehensive study.
\newblock \emph{arXiv preprint arXiv:2510.13681}, 2025.

\bibitem[Dugan et~al.(2023)Dugan, Ippolito, Kirubarajan, Shi, and Callison-Burch]{dugan2023real}
Liam Dugan, Daphne Ippolito, Arun Kirubarajan, Sherry Shi, and Chris Callison-Burch.
\newblock Real or fake text?: Investigating human ability to detect boundaries between human-written and machine-generated text.
\newblock In \emph{Proceedings of the AAAI Conference on Artificial Intelligence}, volume~37, pages 12763--12771, 2023.

\bibitem[Dugan et~al.(2024)Dugan, Hwang, Trhlik, Ludan, Zhu, Xu, Ippolito, and Callison-Burch]{dugan2024raid}
Liam Dugan, Alyssa Hwang, Filip Trhlik, Josh~Magnus Ludan, Andrew Zhu, Hainiu Xu, Daphne Ippolito, and Chris Callison-Burch.
\newblock Raid: A shared benchmark for robust evaluation of machine-generated text detectors.
\newblock \emph{arXiv preprint arXiv:2405.07940}, 2024.

\bibitem[Eger et~al.(2025)Eger, Cao, D'Souza, Geiger, Greisinger, Gross, Hou, Krenn, Lauscher, Li, et~al.]{eger2025transforming}
Steffen Eger, Yong Cao, Jennifer D'Souza, Andreas Geiger, Christian Greisinger, Stephanie Gross, Yufang Hou, Brigitte Krenn, Anne Lauscher, Yizhi Li, et~al.
\newblock Transforming science with large language models: A survey on ai-assisted scientific discovery, experimentation, content generation, and evaluation.
\newblock \emph{arXiv preprint arXiv:2502.05151}, 2025.

\bibitem[Elkhatat et~al.(2023)Elkhatat, Elsaid, and Almeer]{elkhatat2023evaluating}
Ahmed~M Elkhatat, Khaled Elsaid, and Saeed Almeer.
\newblock Evaluating the efficacy of ai content detection tools in differentiating between human and ai-generated text.
\newblock \emph{International Journal for Educational Integrity}, 19\penalty0 (1):\penalty0 17, 2023.

\bibitem[Fang et~al.(2025)Fang, Kong, Zhuang, Qiu, Gao, Chen, Xia, Wang, and Zhang]{fang2025your}
Hao Fang, Jiawei Kong, Tianqu Zhuang, Yixiang Qiu, Kuofeng Gao, Bin Chen, Shu-Tao Xia, Yaowei Wang, and Min Zhang.
\newblock Your language model can secretly write like humans: Contrastive paraphrase attacks on llm-generated text detectors.
\newblock \emph{arXiv preprint arXiv:2505.15337}, 2025.

\bibitem[Fraser et~al.(2024)Fraser, Dawkins, and Kiritchenko]{fraser2024detecting}
Kathleen~C Fraser, Hillary Dawkins, and Svetlana Kiritchenko.
\newblock Detecting ai-generated text: Factors influencing detectability with current methods.
\newblock \emph{arXiv preprint arXiv:2406.15583}, 2024.

\bibitem[Galpin et~al.(2025)Galpin, Anderson, and Juzek]{galpin2025exploring}
Riley Galpin, Bryce Anderson, and Tom~S Juzek.
\newblock Exploring the structure of ai-induced language change in scientific english.
\newblock \emph{arXiv preprint arXiv:2506.21817}, 2025.

\bibitem[Gambini et~al.(2022)Gambini, Fagni, Falchi, and Tesconi]{gambini2022pushing}
Margherita Gambini, Tiziano Fagni, Fabrizio Falchi, and Maurizio Tesconi.
\newblock On pushing deepfake tweet detection capabilities to the limits.
\newblock In \emph{Proceedings of the 14th ACM Web Science Conference 2022}, pages 154--163, 2022.

\bibitem[Gameiro et~al.(2024)Gameiro, Kucharavy, and Dolamic]{gameiro2024llm}
Henrique Da~Silva Gameiro, Andrei Kucharavy, and Ljiljana Dolamic.
\newblock Llm detectors still fall short of real world: Case of llm-generated short news-like posts.
\newblock \emph{arXiv preprint arXiv:2409.03291}, 2024.

\bibitem[Ganie(2025)]{ganie2025uncertainty}
Aadil~Gani Ganie.
\newblock Uncertainty in authorship: Why perfect ai detection is mathematically impossible.
\newblock \emph{arXiv preprint arXiv:2509.11915}, 2025.

\bibitem[Gao et~al.(2025)Gao, Li, Wang, Li, Liu, Song, Zhang, Yan, Nakov, and Chen]{gao2025personalization}
Lang Gao, Xuhui Li, Chenxi Wang, Mingzhe Li, Wei Liu, Zirui Song, Jinghui Zhang, Rui Yan, Preslav Nakov, and Xiuying Chen.
\newblock When personalization tricks detectors: The feature-inversion trap in machine-generated text detection.
\newblock \emph{arXiv preprint arXiv:2510.12476}, 2025.

\bibitem[Gehring and Paa{\ss}en(2025)]{gehring2025assessing}
Lukas Gehring and Benjamin Paa{\ss}en.
\newblock Assessing llm text detection in educational contexts: Does human contribution affect detection?
\newblock \emph{arXiv preprint arXiv:2508.08096}, 2025.

\bibitem[Gehrmann et~al.(2019)Gehrmann, Strobelt, and Rush]{gehrmann2019gltr}
Sebastian Gehrmann, Hendrik Strobelt, and Alexander~M Rush.
\newblock Gltr: Statistical detection and visualization of generated text.
\newblock \emph{arXiv preprint arXiv:1906.04043}, 2019.

\bibitem[Geng and Trotta(2024)]{geng2024chatgpt}
Mingmeng Geng and Roberto Trotta.
\newblock Is chatgpt transforming academics' writing style?
\newblock \emph{arXiv preprint arXiv:2404.08627}, 2024.

\bibitem[Geng and Trotta(2025)]{geng2025human}
Mingmeng Geng and Roberto Trotta.
\newblock Human-llm coevolution: Evidence from academic writing.
\newblock \emph{arXiv preprint arXiv:2502.09606}, 2025.

\bibitem[Geng et~al.(2024)Geng, Chen, Wu, Chen, Wan, and Zhou]{geng2024impact}
Mingmeng Geng, Caixi Chen, Yanru Wu, Dongping Chen, Yao Wan, and Pan Zhou.
\newblock The impact of large language models in academia: from writing to speaking.
\newblock \emph{arXiv preprint arXiv:2409.13686}, 2024.

\bibitem[Ghosal et~al.(2023)Ghosal, Chakraborty, Geiping, Huang, Manocha, and Bedi]{ghosal2023towards}
Soumya~Suvra Ghosal, Souradip Chakraborty, Jonas Geiping, Furong Huang, Dinesh Manocha, and Amrit~Singh Bedi.
\newblock Towards possibilities \& impossibilities of ai-generated text detection: A survey.
\newblock \emph{arXiv preprint arXiv:2310.15264}, 2023.

\bibitem[Giray(2024)]{giray2024problem}
Louie Giray.
\newblock The problem with false positives: Ai detection unfairly accuses scholars of ai plagiarism.
\newblock \emph{The Serials Librarian}, 85\penalty0 (5-6):\penalty0 181--189, 2024.

\bibitem[Grinbaum and Adomaitis(2022)]{grinbaum2022ethical}
Alexei Grinbaum and Laurynas Adomaitis.
\newblock The ethical need for watermarks in machine-generated language.
\newblock \emph{arXiv preprint arXiv:2209.03118}, 2022.

\bibitem[Gritsai et~al.(2024)Gritsai, Voznyuk, Grabovoy, and Chekhovich]{gritsai2024ai}
German Gritsai, Anastasia Voznyuk, Andrey Grabovoy, and Yury Chekhovich.
\newblock Are ai detectors good enough? a survey on quality of datasets with machine-generated texts.
\newblock \emph{arXiv preprint arXiv:2410.14677}, 2024.

\bibitem[Guo et~al.(2025)Guo, Yang, Zhang, Song, Zhang, Xu, Zhu, Ma, Wang, Bi, et~al.]{guo2025deepseek}
Daya Guo, Dejian Yang, Haowei Zhang, Junxiao Song, Ruoyu Zhang, Runxin Xu, Qihao Zhu, Shirong Ma, Peiyi Wang, Xiao Bi, et~al.
\newblock Deepseek-r1: Incentivizing reasoning capability in llms via reinforcement learning.
\newblock \emph{arXiv preprint arXiv:2501.12948}, 2025.

\bibitem[Hadi et~al.(2023)Hadi, Qureshi, Shah, Irfan, Zafar, Shaikh, Akhtar, Wu, Mirjalili, et~al.]{hadi2023survey}
Muhammad~Usman Hadi, Rizwan Qureshi, Abbas Shah, Muhammad Irfan, Anas Zafar, Muhammad~Bilal Shaikh, Naveed Akhtar, Jia Wu, Seyedali Mirjalili, et~al.
\newblock A survey on large language models: Applications, challenges, limitations, and practical usage.
\newblock \emph{Authorea Preprints}, 2023.

\bibitem[Hans et~al.(2024)Hans, Schwarzschild, Cherepanova, Kazemi, Saha, Goldblum, Geiping, and Goldstein]{hans2024spotting}
Abhimanyu Hans, Avi Schwarzschild, Valeriia Cherepanova, Hamid Kazemi, Aniruddha Saha, Micah Goldblum, Jonas Geiping, and Tom Goldstein.
\newblock Spotting llms with binoculars: Zero-shot detection of machine-generated text.
\newblock \emph{arXiv preprint arXiv:2401.12070}, 2024.

\bibitem[He et~al.(2024)He, Shen, Chen, Backes, and Zhang]{he2024mgtbench}
Xinlei He, Xinyue Shen, Zeyuan Chen, Michael Backes, and Yang Zhang.
\newblock Mgtbench: Benchmarking machine-generated text detection.
\newblock In \emph{Proceedings of the 2024 on ACM SIGSAC Conference on Computer and Communications Security}, pages 2251--2265, 2024.

\bibitem[Hendrycks et~al.(2020)Hendrycks, Burns, Basart, Zou, Mazeika, Song, and Steinhardt]{hendrycks2020measuring}
Dan Hendrycks, Collin Burns, Steven Basart, Andy Zou, Mantas Mazeika, Dawn Song, and Jacob Steinhardt.
\newblock Measuring massive multitask language understanding.
\newblock \emph{arXiv preprint arXiv:2009.03300}, 2020.

\bibitem[Hosseini et~al.(2023)Hosseini, Resnik, and Holmes]{hosseini2023ethics}
Mohammad Hosseini, David~B Resnik, and Kristi Holmes.
\newblock The ethics of disclosing the use of artificial intelligence tools in writing scholarly manuscripts.
\newblock \emph{Research Ethics}, 19\penalty0 (4):\penalty0 449--465, 2023.

\bibitem[Hu and Collier(2024)]{hu2024quantifying}
Tiancheng Hu and Nigel Collier.
\newblock Quantifying the persona effect in llm simulations.
\newblock \emph{arXiv preprint arXiv:2402.10811}, 2024.

\bibitem[Huang et~al.(2024)Huang, Zhang, Li, You, Wang, and Yang]{huang2024ai}
Guanhua Huang, Yuchen Zhang, Zhe Li, Yongjian You, Mingze Wang, and Zhouwang Yang.
\newblock Are ai-generated text detectors robust to adversarial perturbations?
\newblock \emph{arXiv preprint arXiv:2406.01179}, 2024.

\bibitem[Huang et~al.(2025)Huang, Xu, Geng, Wan, and Chen]{huang2025wikipedia}
Siming Huang, Yuliang Xu, Mingmeng Geng, Yao Wan, and Dongping Chen.
\newblock Wikipedia in the era of llms: Evolution and risks.
\newblock \emph{arXiv preprint arXiv:2503.02879}, 2025.

\bibitem[Hurst et~al.(2024)Hurst, Lerer, Goucher, Perelman, Ramesh, Clark, Ostrow, Welihinda, Hayes, Radford, et~al.]{hurst2024gpt}
Aaron Hurst, Adam Lerer, Adam~P Goucher, Adam Perelman, Aditya Ramesh, Aidan Clark, AJ~Ostrow, Akila Welihinda, Alan Hayes, Alec Radford, et~al.
\newblock Gpt-4o system card.
\newblock \emph{arXiv preprint arXiv:2410.21276}, 2024.

\bibitem[Iofinova et~al.(2025)Iofinova, Jovanovic, and Alistarh]{iofinova2025position}
Eugenia Iofinova, Andrej Jovanovic, and Dan Alistarh.
\newblock Position: It's time to act on the risk of efficient personalized text generation.
\newblock \emph{arXiv preprint arXiv:2502.06560}, 2025.

\bibitem[Ippolito et~al.(2019)Ippolito, Duckworth, Callison-Burch, and Eck]{ippolito2019automatic}
Daphne Ippolito, Daniel Duckworth, Chris Callison-Burch, and Douglas Eck.
\newblock Automatic detection of generated text is easiest when humans are fooled.
\newblock \emph{arXiv preprint arXiv:1911.00650}, 2019.

\bibitem[Ji et~al.(2024)Ji, Li, Li, Guo, Qiu, Huang, Chen, Jiang, and Lu]{ji2024detecting}
Jiazhou Ji, Ruizhe Li, Shujun Li, Jie Guo, Weidong Qiu, Zheng Huang, Chiyu Chen, Xiaoyu Jiang, and Xinru Lu.
\newblock Detecting machine-generated texts: Not just" ai vs humans" and explainability is complicated.
\newblock \emph{arXiv preprint arXiv:2406.18259}, 2024.

\bibitem[Jovanovi{\'c} et~al.(2024)Jovanovi{\'c}, Staab, and Vechev]{jovanovic2024watermark}
Nikola Jovanovi{\'c}, Robin Staab, and Martin Vechev.
\newblock Watermark stealing in large language models.
\newblock \emph{arXiv preprint arXiv:2402.19361}, 2024.

\bibitem[Juzek and Ward(2024)]{juzek2024does}
Tom~S Juzek and Zina~B Ward.
\newblock Why does chatgpt" delve" so much? exploring the sources of lexical overrepresentation in large language models.
\newblock \emph{arXiv preprint arXiv:2412.11385}, 2024.

\bibitem[Kaddour et~al.(2023)Kaddour, Harris, Mozes, Bradley, Raileanu, and McHardy]{kaddour2023challenges}
Jean Kaddour, Joshua Harris, Maximilian Mozes, Herbie Bradley, Roberta Raileanu, and Robert McHardy.
\newblock Challenges and applications of large language models.
\newblock \emph{arXiv preprint arXiv:2307.10169}, 2023.

\bibitem[Kadiyala et~al.(2025)Kadiyala, Pullakhandam, Mehreen, Sharma, Gupta, Purbey, Srivastava, Tippareddy, Bobbili, Chandrashekhar, et~al.]{kadiyala2025robust}
Ram Mohan~Rao Kadiyala, Siddartha Pullakhandam, Kanwal Mehreen, Drishti Sharma, Siddhant Gupta, Jebish Purbey, Ashay Srivastava, Subhasya Tippareddy, Arvind~Reddy Bobbili, Suraj~Telugara Chandrashekhar, et~al.
\newblock Robust and fine-grained detection of ai generated texts.
\newblock \emph{arXiv preprint arXiv:2504.11952}, 2025.

\bibitem[Kadoma et~al.(2025)Kadoma, Metaxa, and Naaman]{kadoma2025generative}
Kowe Kadoma, Dana{\"e} Metaxa, and Mor Naaman.
\newblock Generative ai and perceptual harms: Who's suspected of using llms?
\newblock In \emph{Proceedings of the 2025 CHI Conference on Human Factors in Computing Systems}, pages 1--17, 2025.

\bibitem[Kar et~al.(2024)Kar, Bansal, Modi, and Singh]{kar2024sensitive}
Sujita~Kumar Kar, Teena Bansal, Sumit Modi, and Amit Singh.
\newblock How sensitive are the free ai-detector tools in detecting ai-generated texts? a comparison of popular ai-detector tools.
\newblock \emph{Indian Journal of Psychological Medicine}, page 02537176241247934, 2024.

\bibitem[Kasneci et~al.(2023)Kasneci, Se{\ss}ler, K{\"u}chemann, Bannert, Dementieva, Fischer, Gasser, Groh, G{\"u}nnemann, H{\"u}llermeier, et~al.]{kasneci2023chatgpt}
Enkelejda Kasneci, Kathrin Se{\ss}ler, Stefan K{\"u}chemann, Maria Bannert, Daryna Dementieva, Frank Fischer, Urs Gasser, Georg Groh, Stephan G{\"u}nnemann, Eyke H{\"u}llermeier, et~al.
\newblock Chatgpt for good? on opportunities and challenges of large language models for education.
\newblock \emph{Learning and individual differences}, 103:\penalty0 102274, 2023.

\bibitem[Kirchenbauer et~al.(2023)Kirchenbauer, Geiping, Wen, Katz, Miers, and Goldstein]{kirchenbauer2023watermark}
John Kirchenbauer, Jonas Geiping, Yuxin Wen, Jonathan Katz, Ian Miers, and Tom Goldstein.
\newblock A watermark for large language models.
\newblock In \emph{International Conference on Machine Learning}, pages 17061--17084. PMLR, 2023.

\bibitem[K{\"o}bis and Mossink(2021)]{kobis2021artificial}
Nils K{\"o}bis and Luca~D Mossink.
\newblock Artificial intelligence versus maya angelou: Experimental evidence that people cannot differentiate ai-generated from human-written poetry.
\newblock \emph{Computers in human behavior}, 114:\penalty0 106553, 2021.

\bibitem[Krishna et~al.(2023)Krishna, Song, Karpinska, Wieting, and Iyyer]{krishna2023paraphrasing}
Kalpesh Krishna, Yixiao Song, Marzena Karpinska, John Wieting, and Mohit Iyyer.
\newblock Paraphrasing evades detectors of ai-generated text, but retrieval is an effective defense.
\newblock \emph{Advances in Neural Information Processing Systems}, 36:\penalty0 27469--27500, 2023.

\bibitem[Kumar et~al.(2025)Kumar, Garg, Sengupta, Ghosal, and Ekbal]{kumar2025mixrevdetect}
Sandeep Kumar, Samarth Garg, Sagnik Sengupta, Tirthankar Ghosal, and Asif Ekbal.
\newblock Mixrevdetect: Towards detecting ai-generated content in hybrid peer reviews.
\newblock In \emph{Proceedings of the 2025 Conference of the Nations of the Americas Chapter of the Association for Computational Linguistics: Human Language Technologies (Volume 2: Short Papers)}, pages 944--953, 2025.

\bibitem[Kumarage et~al.(2023)Kumarage, Garland, Bhattacharjee, Trapeznikov, Ruston, and Liu]{kumarage2023stylometric}
Tharindu Kumarage, Joshua Garland, Amrita Bhattacharjee, Kirill Trapeznikov, Scott Ruston, and Huan Liu.
\newblock Stylometric detection of ai-generated text in twitter timelines.
\newblock \emph{arXiv preprint arXiv:2303.03697}, 2023.

\bibitem[Kumarage et~al.(2024)Kumarage, Agrawal, Sheth, Moraffah, Chadha, Garland, and Liu]{kumarage2024survey}
Tharindu Kumarage, Garima Agrawal, Paras Sheth, Raha Moraffah, Aman Chadha, Joshua Garland, and Huan Liu.
\newblock A survey of ai-generated text forensic systems: Detection, attribution, and characterization.
\newblock \emph{arXiv preprint arXiv:2403.01152}, 2024.

\bibitem[La~Cava et~al.(2025)La~Cava, Aiello, and Tagarelli]{la2025machines}
Lucio La~Cava, Luca~Maria Aiello, and Andrea Tagarelli.
\newblock Machines in the crowd? measuring the footprint of machine-generated text on reddit.
\newblock \emph{arXiv preprint arXiv:2510.07226}, 2025.

\bibitem[Lau and Zubiaga(2024)]{lau2024understanding}
Hiu~Ting Lau and Arkaitz Zubiaga.
\newblock Understanding the effects of human-written paraphrases in llm-generated text detection.
\newblock \emph{arXiv preprint arXiv:2411.03806}, 2024.

\bibitem[Lepp and Smith(2025)]{lepp2025you}
Haley Lepp and Daniel~Scott Smith.
\newblock " you cannot sound like gpt": Signs of language discrimination and resistance in computer science publishing.
\newblock \emph{arXiv preprint arXiv:2505.08127}, 2025.

\bibitem[Lepp{\"a}nen et~al.(2025)Lepp{\"a}nen, Aunimo, Hellas, Nurminen, and Mannila]{leppanen2025large}
Leo Lepp{\"a}nen, Lili Aunimo, Arto Hellas, Jukka~K Nurminen, and Linda Mannila.
\newblock How large language models are changing mooc essay answers: A comparison of pre-and post-llm responses.
\newblock \emph{arXiv preprint arXiv:2504.13038}, 2025.

\bibitem[Li and Wan(2025)]{li2025writes}
Jiatao Li and Xiaojun Wan.
\newblock Who writes what: Unveiling the impact of author roles on ai-generated text detection.
\newblock \emph{arXiv preprint arXiv:2502.12611}, 2025.

\bibitem[Liang et~al.(2023)Liang, Yuksekgonul, Mao, Wu, and Zou]{liang2023gpt}
Weixin Liang, Mert Yuksekgonul, Yining Mao, Eric Wu, and James Zou.
\newblock Gpt detectors are biased against non-native english writers.
\newblock \emph{Patterns}, 4\penalty0 (7), 2023.

\bibitem[Liang et~al.(2024)Liang, Izzo, Zhang, Lepp, Cao, Zhao, Chen, Ye, Liu, Huang, et~al.]{liang2024monitoring}
Weixin Liang, Zachary Izzo, Yaohui Zhang, Haley Lepp, Hancheng Cao, Xuandong Zhao, Lingjiao Chen, Haotian Ye, Sheng Liu, Zhi Huang, et~al.
\newblock Monitoring ai-modified content at scale: A case study on the impact of chatgpt on ai conference peer reviews.
\newblock \emph{arXiv preprint arXiv:2403.07183}, 2024.

\bibitem[Liang et~al.(2025)Liang, Zhang, Codreanu, Wang, Cao, and Zou]{liang2025widespread}
Weixin Liang, Yaohui Zhang, Mihai Codreanu, Jiayu Wang, Hancheng Cao, and James Zou.
\newblock The widespread adoption of large language model-assisted writing across society.
\newblock \emph{arXiv preprint arXiv:2502.09747}, 2025.

\bibitem[Liu et~al.(2024)Liu, Feng, Xue, Wang, Wu, Lu, Zhao, Deng, Zhang, Ruan, et~al.]{liu2024deepseek}
Aixin Liu, Bei Feng, Bing Xue, Bingxuan Wang, Bochao Wu, Chengda Lu, Chenggang Zhao, Chengqi Deng, Chenyu Zhang, Chong Ruan, et~al.
\newblock Deepseek-v3 technical report.
\newblock \emph{arXiv preprint arXiv:2412.19437}, 2024.

\bibitem[Liu et~al.(2023)Liu, Huang, Yu, Zhang, Wu, Cao, Dai, Zhao, Li, Shu, et~al.]{liu2023deid}
Zhengliang Liu, Yue Huang, Xiaowei Yu, Lu~Zhang, Zihao Wu, Chao Cao, Haixing Dai, Lin Zhao, Yiwei Li, Peng Shu, et~al.
\newblock Deid-gpt: Zero-shot medical text de-identification by gpt-4.
\newblock \emph{arXiv preprint arXiv:2303.11032}, 2023.

\bibitem[Liyanage et~al.(2022)Liyanage, Buscaldi, and Nazarenko]{liyanage2022benchmark}
Vijini Liyanage, Davide Buscaldi, and Adeline Nazarenko.
\newblock A benchmark corpus for the detection of automatically generated text in academic publications.
\newblock \emph{arXiv preprint arXiv:2202.02013}, 2022.

\bibitem[Ma and Wang(2024)]{ma2024zero}
Shixuan Ma and Quan Wang.
\newblock Zero-shot detection of llm-generated text using token cohesiveness.
\newblock \emph{arXiv preprint arXiv:2409.16914}, 2024.

\bibitem[Macko et~al.(2024)Macko, Moro, Uchendu, Srba, Lucas, Yamashita, Tripto, Lee, Simko, and Bielikova]{macko2024authorship}
Dominik Macko, Robert Moro, Adaku Uchendu, Ivan Srba, Jason~Samuel Lucas, Michiharu Yamashita, Nafis~Irtiza Tripto, Dongwon Lee, Jakub Simko, and Maria Bielikova.
\newblock Authorship obfuscation in multilingual machine-generated text detection.
\newblock \emph{arXiv preprint arXiv:2401.07867}, 2024.

\bibitem[McGovern et~al.(2024)McGovern, Stureborg, Suhara, and Alikaniotis]{mcgovern2024your}
Hope McGovern, Rickard Stureborg, Yoshi Suhara, and Dimitris Alikaniotis.
\newblock Your large language models are leaving fingerprints.
\newblock \emph{arXiv preprint arXiv:2405.14057}, 2024.

\bibitem[Mili{\v{c}}ka et~al.(2025)Mili{\v{c}}ka, Marklov{\'a}, Drobil, and Posp{\'\i}{\v{s}}ilov{\'a}]{milivcka2025humans}
Ji{\v{r}}{\'\i} Mili{\v{c}}ka, Anna Marklov{\'a}, Ond{\v{r}}ej Drobil, and Eva Posp{\'\i}{\v{s}}ilov{\'a}.
\newblock Humans can learn to detect ai-generated texts, or at least learn when they can't.
\newblock \emph{arXiv preprint arXiv:2505.01877}, 2025.

\bibitem[Mitchell et~al.(2023)Mitchell, Lee, Khazatsky, Manning, and Finn]{mitchell2023detectgpt}
Eric Mitchell, Yoonho Lee, Alexander Khazatsky, Christopher~D Manning, and Chelsea Finn.
\newblock Detectgpt: Zero-shot machine-generated text detection using probability curvature.
\newblock In \emph{International Conference on Machine Learning}, pages 24950--24962. PMLR, 2023.

\bibitem[Mu{\~n}oz-Ortiz et~al.(2024)Mu{\~n}oz-Ortiz, G{\'o}mez-Rodr{\'\i}guez, and Vilares]{munoz2024contrasting}
Alberto Mu{\~n}oz-Ortiz, Carlos G{\'o}mez-Rodr{\'\i}guez, and David Vilares.
\newblock Contrasting linguistic patterns in human and llm-generated news text.
\newblock \emph{Artificial Intelligence Review}, 57\penalty0 (10):\penalty0 265, 2024.

\bibitem[Nicks et~al.(2023)Nicks, Mitchell, Rafailov, Sharma, Manning, Finn, and Ermon]{nicks2023language}
Charlotte Nicks, Eric Mitchell, Rafael Rafailov, Archit Sharma, Christopher~D Manning, Chelsea Finn, and Stefano Ermon.
\newblock Language model detectors are easily optimized against.
\newblock In \emph{The twelfth international conference on learning representations}, 2023.

\bibitem[Noy and Zhang(2023)]{noy2023experimental}
Shakked Noy and Whitney Zhang.
\newblock Experimental evidence on the productivity effects of generative artificial intelligence.
\newblock \emph{Science}, 381\penalty0 (6654):\penalty0 187--192, 2023.

\bibitem[Pedreschi et~al.(2024)Pedreschi, Pappalardo, Ferragina, Baeza-Yates, Barab{\'a}si, Dignum, Dignum, Eliassi-Rad, Giannotti, Kert{\'e}sz, et~al.]{pedreschi2024human}
Dino Pedreschi, Luca Pappalardo, Emanuele Ferragina, Ricardo Baeza-Yates, Albert-L{\'a}szl{\'o} Barab{\'a}si, Frank Dignum, Virginia Dignum, Tina Eliassi-Rad, Fosca Giannotti, J{\'a}nos Kert{\'e}sz, et~al.
\newblock Human-ai coevolution.
\newblock \emph{Artificial Intelligence}, page 104244, 2024.

\bibitem[Pedrotti et~al.(2025)Pedrotti, Papucci, Ciaccio, Miaschi, Puccetti, Dell'Orletta, and Esuli]{pedrotti2025stress}
Andrea Pedrotti, Michele Papucci, Cristiano Ciaccio, Alessio Miaschi, Giovanni Puccetti, Felice Dell'Orletta, and Andrea Esuli.
\newblock Stress-testing machine generated text detection: Shifting language models writing style to fool detectors.
\newblock \emph{arXiv preprint arXiv:2505.24523}, 2025.

\bibitem[Pu and Demberg(2023)]{pu2023chatgpt}
Dongqi Pu and Vera Demberg.
\newblock Chatgpt vs human-authored text: Insights into controllable text summarization and sentence style transfer.
\newblock \emph{arXiv preprint arXiv:2306.07799}, 2023.

\bibitem[Pudasaini et~al.(2024)Pudasaini, Miralles-Pechu{\'a}n, Lillis, and Salvador]{pudasaini2024survey}
Shushanta Pudasaini, Luis Miralles-Pechu{\'a}n, David Lillis, and Marisa~Llorens Salvador.
\newblock Survey on plagiarism detection in large language models: The impact of chatgpt and gemini on academic integrity.
\newblock \emph{arXiv preprint arXiv:2407.13105}, 2024.

\bibitem[Pudasaini et~al.(2025)Pudasaini, Miralles, Lillis, and Salvador]{pudasaini2025benchmarking}
Shushanta Pudasaini, Luis Miralles, David Lillis, and Marisa~Llorens Salvador.
\newblock Benchmarking ai text detection: Assessing detectors against new datasets, evasion tactics, and enhanced llms.
\newblock In \emph{Proceedings of the 1stWorkshop on GenAI Content Detection (GenAIDetect)}, pages 68--77, 2025.

\bibitem[Quaremba et~al.(2025)Quaremba, Black, Vrande{\v{c}}i{\'c}, and Simperl]{quaremba2025wetbench}
Gerrit Quaremba, Elizabeth Black, Denny Vrande{\v{c}}i{\'c}, and Elena Simperl.
\newblock Wetbench: A benchmark for detecting task-specific machine-generated text on wikipedia.
\newblock \emph{arXiv preprint arXiv:2507.03373}, 2025.

\bibitem[Radford et~al.(2019)Radford, Wu, Child, Luan, Amodei, Sutskever, et~al.]{radford2019language}
Alec Radford, Jeffrey Wu, Rewon Child, David Luan, Dario Amodei, Ilya Sutskever, et~al.
\newblock Language models are unsupervised multitask learners.
\newblock \emph{OpenAI blog}, 1\penalty0 (8):\penalty0 9, 2019.

\bibitem[Rashidi et~al.(2023)Rashidi, Fennell, Albahra, Hu, and Gorbett]{rashidi2023chatgpt}
Hooman~H Rashidi, Brandon~D Fennell, Samer Albahra, Bo~Hu, and Tom Gorbett.
\newblock The chatgpt conundrum: Human-generated scientific manuscripts misidentified as ai creations by ai text detection tool.
\newblock \emph{Journal of Pathology Informatics}, 14:\penalty0 100342, 2023.

\bibitem[Reinhart et~al.(2025)Reinhart, Markey, Laudenbach, Pantusen, Yurko, Weinberg, and Brown]{reinhart2025llms}
Alex Reinhart, Ben Markey, Michael Laudenbach, Kachatad Pantusen, Ronald Yurko, Gordon Weinberg, and David~West Brown.
\newblock Do llms write like humans? variation in grammatical and rhetorical styles.
\newblock \emph{Proceedings of the National Academy of Sciences}, 122\penalty0 (8):\penalty0 e2422455122, 2025.

\bibitem[Rosenfeld and Lazebnik(2024)]{rosenfeld2024whose}
Ariel Rosenfeld and Teddy Lazebnik.
\newblock Whose llm is it anyway? linguistic comparison and llm attribution for gpt-3.5, gpt-4 and bard.
\newblock \emph{arXiv preprint arXiv:2402.14533}, 2024.

\bibitem[Russell et~al.(2025)Russell, Karpinska, and Iyyer]{russell2025people}
Jenna Russell, Marzena Karpinska, and Mohit Iyyer.
\newblock People who frequently use chatgpt for writing tasks are accurate and robust detectors of ai-generated text.
\newblock \emph{arXiv preprint arXiv:2501.15654}, 2025.

\bibitem[Sadasivan et~al.(2023)Sadasivan, Kumar, Balasubramanian, Wang, and Feizi]{sadasivan2023can}
Vinu~Sankar Sadasivan, Aounon Kumar, Sriram Balasubramanian, Wenxiao Wang, and Soheil Feizi.
\newblock Can ai-generated text be reliably detected?
\newblock \emph{arXiv preprint arXiv:2303.11156}, 2023.

\bibitem[Saha and Feizi(2025)]{saha2025almost}
Shoumik Saha and Soheil Feizi.
\newblock Almost ai, almost human: The challenge of detecting ai-polished writing.
\newblock \emph{arXiv preprint arXiv:2502.15666}, 2025.

\bibitem[Sarkar(2025)]{sarkar2025ai}
Advait Sarkar.
\newblock Ai could have written this: Birth of a classist slur in knowledge work.
\newblock In \emph{Proceedings of the Extended Abstracts of the CHI Conference on Human Factors in Computing Systems}, pages 1--12, 2025.

\bibitem[Schuster et~al.(2020)Schuster, Schuster, Shah, and Barzilay]{schuster2020limitations}
Tal Schuster, Roei Schuster, Darsh~J Shah, and Regina Barzilay.
\newblock The limitations of stylometry for detecting machine-generated fake news.
\newblock \emph{Computational Linguistics}, 46\penalty0 (2):\penalty0 499--510, 2020.

\bibitem[Shi et~al.(2024)Shi, Sheng, Cao, Mi, Hu, and Wang]{shi2024ten}
Yuhui Shi, Qiang Sheng, Juan Cao, Hao Mi, Beizhe Hu, and Danding Wang.
\newblock Ten words only still help: Improving black-box ai-generated text detection via proxy-guided efficient re-sampling.
\newblock \emph{arXiv preprint arXiv:2402.09199}, 2024.

\bibitem[Solaiman et~al.(2019)Solaiman, Brundage, Clark, Askell, Herbert-Voss, Wu, Radford, Krueger, Kim, Kreps, et~al.]{solaiman2019release}
Irene Solaiman, Miles Brundage, Jack Clark, Amanda Askell, Ariel Herbert-Voss, Jeff Wu, Alec Radford, Gretchen Krueger, Jong~Wook Kim, Sarah Kreps, et~al.
\newblock Release strategies and the social impacts of language models.
\newblock \emph{arXiv preprint arXiv:1908.09203}, 2019.

\bibitem[Srivastava et~al.(2022)Srivastava, Rastogi, Rao, Shoeb, Abid, Fisch, Brown, Santoro, Gupta, Garriga-Alonso, et~al.]{srivastava2022beyond}
Aarohi Srivastava, Abhinav Rastogi, Abhishek Rao, Abu Awal~Md Shoeb, Abubakar Abid, Adam Fisch, Adam~R Brown, Adam Santoro, Aditya Gupta, Adri{\`a} Garriga-Alonso, et~al.
\newblock Beyond the imitation game: Quantifying and extrapolating the capabilities of language models.
\newblock \emph{arXiv preprint arXiv:2206.04615}, 2022.

\bibitem[Su et~al.(2023)Su, Zhuo, Wang, and Nakov]{su2023detectllm}
Jinyan Su, Terry~Yue Zhuo, Di~Wang, and Preslav Nakov.
\newblock Detectllm: Leveraging log rank information for zero-shot detection of machine-generated text.
\newblock \emph{arXiv preprint arXiv:2306.05540}, 2023.

\bibitem[Sun et~al.(2025)Sun, Yin, Xu, Kolter, and Liu]{sun2025idiosyncrasies}
Mingjie Sun, Yida Yin, Zhiqiu Xu, J~Zico Kolter, and Zhuang Liu.
\newblock Idiosyncrasies in large language models.
\newblock \emph{arXiv preprint arXiv:2502.12150}, 2025.

\bibitem[Sun et~al.(2024)Sun, Zhang, Shen, Zhang, Liu, Backes, Zhang, and He]{sun2024we}
Zhen Sun, Zongmin Zhang, Xinyue Shen, Ziyi Zhang, Yule Liu, Michael Backes, Yang Zhang, and Xinlei He.
\newblock Are we in the ai-generated text world already? quantifying and monitoring aigt on social media.
\newblock \emph{arXiv preprint arXiv:2412.18148}, 2024.

\bibitem[Tang et~al.(2024)Tang, Chuang, and Hu]{tang2024science}
Ruixiang Tang, Yu-Neng Chuang, and Xia Hu.
\newblock The science of detecting llm-generated text.
\newblock \emph{Communications of the ACM}, 67\penalty0 (4):\penalty0 50--59, 2024.

\bibitem[Tao et~al.(2024)Tao, Li, Xi, and Xu]{tao2024cudrt}
Zhen Tao, Zhiyu Li, Dinghao Xi, and Wei Xu.
\newblock Cudrt: Benchmarking the detection of human vs. large language models generated texts.
\newblock \emph{arXiv preprint arXiv:2406.09056}, 2024.

\bibitem[Tufts et~al.(2024)Tufts, Zhao, and Li]{tufts2024practical}
Brian Tufts, Xuandong Zhao, and Lei Li.
\newblock A practical examination of ai-generated text detectors for large language models.
\newblock \emph{arXiv preprint arXiv:2412.05139}, 2024.

\bibitem[Turing(2007)]{turing2007computing}
Alan~M Turing.
\newblock Computing machinery and intelligence.
\newblock In \emph{Parsing the Turing test: Philosophical and methodological issues in the quest for the thinking computer}, pages 23--65. Springer, 2007.

\bibitem[Wahle et~al.(2022)Wahle, Ruas, Kirstein, and Gipp]{wahle2022large}
Jan~Philip Wahle, Terry Ruas, Frederic Kirstein, and Bela Gipp.
\newblock How large language models are transforming machine-paraphrased plagiarism.
\newblock \emph{arXiv preprint arXiv:2210.03568}, 2022.

\bibitem[Walters(2023)]{walters2023effectiveness}
William~H Walters.
\newblock The effectiveness of software designed to detect ai-generated writing: A comparison of 16 ai text detectors.
\newblock \emph{Open Information Science}, 7\penalty0 (1):\penalty0 20220158, 2023.

\bibitem[Wang et~al.(2018)Wang, Singh, Michael, Hill, Levy, and Bowman]{wang2018glue}
Alex Wang, Amanpreet Singh, Julian Michael, Felix Hill, Omer Levy, and Samuel~R Bowman.
\newblock Glue: A multi-task benchmark and analysis platform for natural language understanding.
\newblock \emph{arXiv preprint arXiv:1804.07461}, 2018.

\bibitem[Wang and Li(2025)]{wang2025continually}
Quan Wang and Haoran Li.
\newblock On continually tracing origins of llm-generated text and its application in detecting cheating in student coursework.
\newblock \emph{Big Data and Cognitive Computing}, 9\penalty0 (3):\penalty0 50, 2025.

\bibitem[Wang et~al.(2024{\natexlab{a}})Wang, Chen, Zhou, Ma, Duan, Kang, Yang, Fan, and Tan]{wang2024llm}
Rongsheng Wang, Haoming Chen, Ruizhe Zhou, Han Ma, Yaofei Duan, Yanlan Kang, Songhua Yang, Baoyu Fan, and Tao Tan.
\newblock Llm-detector: Improving ai-generated chinese text detection with open-source llm instruction tuning.
\newblock \emph{arXiv preprint arXiv:2402.01158}, 2024{\natexlab{a}}.

\bibitem[Wang et~al.(2024{\natexlab{b}})Wang, Mansurov, Ivanov, Su, Shelmanov, Tsvigun, Afzal, Mahmoud, Puccetti, Arnold, et~al.]{wang2024m4gt}
Yuxia Wang, Jonibek Mansurov, Petar Ivanov, Jinyan Su, Artem Shelmanov, Akim Tsvigun, Osama~Mohanned Afzal, Tarek Mahmoud, Giovanni Puccetti, Thomas Arnold, et~al.
\newblock M4gt-bench: Evaluation benchmark for black-box machine-generated text detection.
\newblock \emph{arXiv preprint arXiv:2402.11175}, 2024{\natexlab{b}}.

\bibitem[Wang et~al.(2025)Wang, Shelmanov, Mansurov, Tsvigun, Mikhailov, Xing, Xie, Geng, Puccetti, Artemova, et~al.]{wang2025genai}
Yuxia Wang, Artem Shelmanov, Jonibek Mansurov, Akim Tsvigun, Vladislav Mikhailov, Rui Xing, Zhuohan Xie, Jiahui Geng, Giovanni Puccetti, Ekaterina Artemova, et~al.
\newblock Genai content detection task 1: English and multilingual machine-generated text detection: Ai vs. human.
\newblock \emph{arXiv preprint arXiv:2501.11012}, 2025.

\bibitem[Weber-Wulff et~al.(2023)Weber-Wulff, Anohina-Naumeca, Bjelobaba, Folt{\`y}nek, Guerrero-Dib, Popoola, {\v{S}}igut, and Waddington]{weber2023testing}
Debora Weber-Wulff, Alla Anohina-Naumeca, Sonja Bjelobaba, Tom{\'a}{\v{s}} Folt{\`y}nek, Jean Guerrero-Dib, Olumide Popoola, Petr {\v{S}}igut, and Lorna Waddington.
\newblock Testing of detection tools for ai-generated text.
\newblock \emph{International Journal for Educational Integrity}, 19\penalty0 (1):\penalty0 1--39, 2023.

\bibitem[Wu et~al.(2024)Wu, Zhan, Wong, Yang, Yang, Yuan, and Chao]{wu2024detectrl}
Junchao Wu, Runzhe Zhan, Derek Wong, Shu Yang, Xinyi Yang, Yulin Yuan, and Lidia Chao.
\newblock Detectrl: Benchmarking llm-generated text detection in real-world scenarios.
\newblock \emph{Advances in Neural Information Processing Systems}, 37:\penalty0 100369--100401, 2024.

\bibitem[Wu et~al.(2025)Wu, Yang, Zhan, Yuan, Chao, and Wong]{wu2025survey}
Junchao Wu, Shu Yang, Runzhe Zhan, Yulin Yuan, Lidia~Sam Chao, and Derek~Fai Wong.
\newblock A survey on llm-generated text detection: Necessity, methods, and future directions.
\newblock \emph{Computational Linguistics}, pages 1--66, 2025.

\bibitem[Wu et~al.(2023)Wu, Pang, Shen, Cheng, and Chua]{wu2023llmdet}
Kangxi Wu, Liang Pang, Huawei Shen, Xueqi Cheng, and Tat-Seng Chua.
\newblock Llmdet: A third party large language models generated text detection tool.
\newblock \emph{arXiv preprint arXiv:2305.15004}, 2023.

\bibitem[Xie et~al.(2025)Xie, Siddeek, Seif, Goldsmith, and Wang]{xie2025detecting}
Liyan Xie, Muhammad Siddeek, Mohamed Seif, Andrea~J Goldsmith, and Mengdi Wang.
\newblock Detecting post-generation edits to watermarked llm outputs via combinatorial watermarking.
\newblock \emph{arXiv preprint arXiv:2510.01637}, 2025.

\bibitem[Yakura et~al.(2024)Yakura, Lopez-Lopez, Brinkmann, Serna, Gupta, and Rahwan]{yakura2024empirical}
Hiromu Yakura, Ezequiel Lopez-Lopez, Levin Brinkmann, Ignacio Serna, Prateek Gupta, and Iyad Rahwan.
\newblock Empirical evidence of large language model's influence on human spoken communication.
\newblock \emph{arXiv preprint arXiv:2409.01754}, 2024.

\bibitem[Yang et~al.(2023)Yang, Pan, Zhao, Chen, Petzold, Wang, and Cheng]{yang2023survey}
Xianjun Yang, Liangming Pan, Xuandong Zhao, Haifeng Chen, Linda Petzold, William~Yang Wang, and Wei Cheng.
\newblock A survey on detection of llms-generated content.
\newblock \emph{arXiv preprint arXiv:2310.15654}, 2023.

\bibitem[Ye et~al.(2025)Ye, Wang, Zou, Yan, Wang, Zheng, Xu, King, Yu, and Wen]{ye2025position}
Jingheng Ye, Shen Wang, Deqing Zou, Yibo Yan, Kun Wang, Hai-Tao Zheng, Zenglin Xu, Irwin King, Philip~S Yu, and Qingsong Wen.
\newblock Position: Llms can be good tutors in foreign language education.
\newblock \emph{arXiv preprint arXiv:2502.05467}, 2025.

\bibitem[Yu et~al.(2024{\natexlab{a}})Yu, Chen, Yang, Zhang, and Yu]{yu2024text}
Xiao Yu, Kejiang Chen, Qi~Yang, Weiming Zhang, and Nenghai Yu.
\newblock Text fluoroscopy: Detecting llm-generated text through intrinsic features.
\newblock In \emph{Proceedings of the 2024 Conference on Empirical Methods in Natural Language Processing}, pages 15838--15846, 2024{\natexlab{a}}.

\bibitem[Yu et~al.(2024{\natexlab{b}})Yu, Qi, Chen, Chen, Yang, Zhu, Shang, Zhang, and Yu]{yu2024dpic}
Xiao Yu, Yuang Qi, Kejiang Chen, Guoqiang Chen, Xi~Yang, Pengyuan Zhu, Xiuwei Shang, Weiming Zhang, and Nenghai Yu.
\newblock Dpic: Decoupling prompt and intrinsic characteristics for llm generated text detection.
\newblock \emph{Advances in Neural Information Processing Systems}, 37:\penalty0 16194--16212, 2024{\natexlab{b}}.

\bibitem[Zaitsu and Jin(2023)]{zaitsu2023distinguishing}
Wataru Zaitsu and Mingzhe Jin.
\newblock Distinguishing chatgpt (-3.5,-4)-generated and human-written papers through japanese stylometric analysis.
\newblock \emph{PLoS One}, 18\penalty0 (8):\penalty0 e0288453, 2023.

\bibitem[Zellers et~al.(2019)Zellers, Holtzman, Rashkin, Bisk, Farhadi, Roesner, and Choi]{zellers2019defending}
Rowan Zellers, Ari Holtzman, Hannah Rashkin, Yonatan Bisk, Ali Farhadi, Franziska Roesner, and Yejin Choi.
\newblock Defending against neural fake news.
\newblock \emph{Advances in neural information processing systems}, 32, 2019.

\bibitem[Zeng et~al.(2025)Zeng, Tang, Chen, Shen, Yu, Zhao, Chen, Cheng, and Xu]{zeng2025human}
Cong Zeng, Shengkun Tang, Yuanzhou Chen, Zhiqiang Shen, Wenchao Yu, Xujiang Zhao, Haifeng Chen, Wei Cheng, and Zhiqiang Xu.
\newblock Human texts are outliers: Detecting llm-generated texts via out-of-distribution detection.
\newblock \emph{arXiv preprint arXiv:2510.08602}, 2025.

\bibitem[Zhang et~al.(2024{\natexlab{a}})Zhang, Gao, Chen, Huang, Huang, Sun, Zhang, Li, Fu, Wan, et~al.]{zhang2024llm}
Qihui Zhang, Chujie Gao, Dongping Chen, Yue Huang, Yixin Huang, Zhenyang Sun, Shilin Zhang, Weiye Li, Zhengyan Fu, Yao Wan, et~al.
\newblock Llm-as-a-coauthor: Can mixed human-written and machine-generated text be detected?
\newblock \emph{arXiv preprint arXiv:2401.05952}, 2024{\natexlab{a}}.

\bibitem[Zhang et~al.(2024{\natexlab{b}})Zhang, Ma, Liu, Liu, Wang, and Lu]{zhang2024detection}
Yuehan Zhang, Yongqiang Ma, Jiawei Liu, Xiaozhong Liu, Xiaofeng Wang, and Wei Lu.
\newblock Detection vs. anti-detection: Is text generated by ai detectable?
\newblock In \emph{International Conference on Information}, pages 209--222. Springer, 2024{\natexlab{b}}.

\bibitem[Zhang et~al.(2025)Zhang, Zhang, Zhang, Zhang, Pan, Liu, Gill, and Zhang]{zhang2025character}
Zhaoxi Zhang, Xiaomei Zhang, Yanjun Zhang, He~Zhang, Shirui Pan, Bo~Liu, Asif~Qumer Gill, and Leo~Yu Zhang.
\newblock Character-level perturbations disrupt llm watermarks.
\newblock \emph{arXiv preprint arXiv:2509.09112}, 2025.

\bibitem[Zhou et~al.(2025)Zhou, Zhu, Su, Ye, Yang, Gavioli-Akilagun, and Shi]{zhou2025adadetectgpt}
Hongyi Zhou, Jin Zhu, Pingfan Su, Kai Ye, Ying Yang, Shakeel~AOB Gavioli-Akilagun, and Chengchun Shi.
\newblock Adadetectgpt: Adaptive detection of llm-generated text with statistical guarantees.
\newblock \emph{arXiv preprint arXiv:2510.01268}, 2025.

\bibitem[Zhou et~al.(2024)Zhou, He, and Sun]{zhou2024humanizing}
Ying Zhou, Ben He, and Le~Sun.
\newblock Humanizing machine-generated content: evading ai-text detection through adversarial attack.
\newblock \emph{arXiv preprint arXiv:2404.01907}, 2024.

\bibitem[Zhu et~al.(2023)Zhu, Yuan, Cui, Chen, Fu, He, Deng, Liu, Sun, and Gu]{zhu2023beat}
Biru Zhu, Lifan Yuan, Ganqu Cui, Yangyi Chen, Chong Fu, Bingxiang He, Yangdong Deng, Zhiyuan Liu, Maosong Sun, and Ming Gu.
\newblock Beat llms at their own game: Zero-shot llm-generated text detection via querying chatgpt.
\newblock In \emph{Proceedings of the 2023 Conference on Empirical Methods in Natural Language Processing}, pages 7470--7483, 2023.

\bibitem[Zhu and Jurgens(2021)]{zhu2021idiosyncratic}
Jian Zhu and David Jurgens.
\newblock Idiosyncratic but not arbitrary: Learning idiolects in online registers reveals distinctive yet consistent individual styles.
\newblock \emph{arXiv preprint arXiv:2109.03158}, 2021.

\end{thebibliography}

%%%%%%%%%%%%%%%%%%%%%%%%%%%%%%%%%%%%%%%%%%%%%%%%%%%%%%%%%%%%

\newpage

\appendix

\section{Simulation Results}
\label{simulation_results}

The temperature parameter is set to 0 in the simulations.

\begin{itemize}
\item \textbf{GPT-3.5-turbo \& Polish the following passage (provide only the result): }I propose considering the question, "Can machines think?" This should start with defining the terms "machine" and "think." Definitions should reflect the common use of the words, but this approach is risky. If we try to find the meaning of "machine" and "think" by looking at how they are commonly used, it seems that the answer to the question must be sought in a statistical survey like a Gallup poll. However, this is absurd. Instead of trying to define it in this way, I will replace the question with another closely related one, expressed in clearer terms.
\item \textbf{GPT-4o-mini \& Polish the following passage (provide only the result): }I propose to explore the question, "Can machines think?" To begin, we must define the terms "machine" and "think." While these definitions could be framed to reflect their common usage, this approach is fraught with danger. Relying solely on how these words are typically used leads to the troubling conclusion that the meaning of "machine" and "think," as well as the answer to the question, "Can machines think?" could be determined through a statistical survey, such as a Gallup poll. This notion is absurd. Rather than pursuing such definitions, I will rephrase the question into a closely related one that is expressed in clearer terms.
\item \textbf{GPT-4o \& Polish the following passage (provide only the result): }I propose to consider the question, "Can machines think?" This should begin with defining the terms "machine" and "think." While these definitions might aim to reflect the common usage of the words, this approach is risky. If we determine the meanings of "machine" and "think" based on their everyday use, we might conclude that the answer to "Can machines think?" should be derived from a statistical survey, like a Gallup poll. However, this is absurd. Instead of attempting such a definition, I will replace the question with another, closely related one, expressed in relatively unambiguous terms.
\item \textbf{GPT-3.5-turbo \& Rewrite the following passage (provide only the result):}The question of whether machines can think should not be based on common usage of the terms "machine" and "think." Instead, I will propose a different question that is closely related and more clearly defined.
\item \textbf{GPT-4o-mini \& Rewrite the following passage (provide only the result):}I propose to explore the question, "Can machines think?" This exploration should start with defining the terms "machine" and "think." While these definitions could aim to reflect their typical usage, this approach is fraught with risks. Relying on common usage to determine the meanings of "machine" and "think" leads to the troubling conclusion that the answer to "Can machines think?" might be found through a statistical survey, like a Gallup poll. This notion is absurd. Rather than pursuing such definitions, I will reframe the question with another that is closely related and articulated in clearer terms.
\item \textbf{GPT-4o \& Rewrite the following passage (provide only the result):}I propose to explore the question, "Can machines think?" by first defining the terms "machine" and "think." While these definitions should ideally reflect the common usage of the words, this approach is risky. If we determine the meanings of "machine" and "think" based on their typical usage, we might conclude that the answer to "Can machines think?" should be derived from a statistical survey like a Gallup poll, which is unreasonable. Therefore, instead of trying to define these terms, I will reframe the question into another that is closely related and expressed in clearer terms.
\item \textbf{GPT-3.5-turbo \& Please polish the following passage (provide only the result): }I propose to explore the question, "Can machines think?" This exploration will begin with defining the terms "machine" and "think." While it may be tempting to rely on common usage for these definitions, this approach is risky. If we simply look at how these words are commonly used, we may be led to the mistaken belief that the answer to the question lies in a statistical survey like a Gallup poll. However, this is clearly illogical. Rather than pursuing such a definition, I will instead pose a related question that is more clearly defined and easier to address.
\item \textbf{GPT-4o-mini \& Please polish the following passage (provide only the result): }I propose to explore the question, "Can machines think?" This inquiry should begin with clear definitions of the terms "machine" and "think." While these definitions could be framed to reflect common usage, this approach is fraught with challenges. Relying solely on how these words are typically used may lead to the conclusion that the answer to "Can machines think?" can be determined through a statistical survey, such as a Gallup poll. However, this notion is absurd. Rather than pursuing such definitions, I will rephrase the question into a closely related one that is expressed in more precise terms.
\item \textbf{GPT-4o \& Please polish the following passage (provide only the result): }I propose to consider the question, "Can machines think?" This should begin with defining the terms "machine" and "think." While these definitions might aim to reflect the common usage of the words, this approach is risky. If we determine the meanings of "machine" and "think" by examining their everyday use, we might conclude that the answer to "Can machines think?" should be sought through a statistical survey, like a Gallup poll. However, this is absurd. Instead of attempting such a definition, I will replace the question with another, closely related one, expressed in relatively unambiguous terms.
\item \textbf{GPT-3.5-turbo \& Please rewrite the following passage (provide only the result): }The question of whether machines can think can be reframed as a different question that is more clear and closely related.
\item \textbf{GPT-4o-mini \& Please rewrite the following passage (provide only the result): }I propose to explore the question, "Can machines think?" This exploration should start with defining the terms "machine" and "think." While these definitions could aim to reflect their typical usage, this approach is fraught with risks. Relying on common usage to determine the meanings of "machine" and "think" leads to the troubling conclusion that the answer to "Can machines think?" might be found through a statistical survey, like a Gallup poll. This notion is absurd. Rather than pursuing such definitions, I will reframe the question with another that is closely related and articulated in clearer terms.
\item \textbf{GPT-4o \& Please rewrite the following passage (provide only the result): }I suggest we examine the question, "Can machines think?" by first defining the terms "machine" and "think." While these definitions should ideally reflect common usage, relying solely on this approach is risky. If we determine the meanings of "machine" and "think" based on their typical usage, we might conclude that the answer to "Can machines think?" should be derived from a statistical survey like a Gallup poll, which is clearly unreasonable. Therefore, rather than attempting such a definition, I will substitute the question with another one that is closely related and expressed in clearer terms.
\item \textbf{DeepSeek-V3.2 \& Polish the following passage (provide only the result): }I propose to consider the question, "Can machines think?" To do so, one might begin by defining the terms "machine" and "think." These definitions could be framed to reflect the normal use of the words as far as possible, but this approach is problematic. If we seek the meaning of "machine" and "think" by examining common usage, we are led to the conclusion that the meaning—and the answer to the question—must be determined by a statistical survey, such as a Gallup poll. This, however, is absurd. Rather than attempting such a definition, I will replace the original question with another, closely related to it and expressed in relatively unambiguous words.
\item \textbf{DeepSeek-R1 \& Polish the following passage (provide only the result): }I propose to consider the question, "Can machines think?" This should begin by defining the terms "machine" and "think." One might frame these definitions to reflect the normal use of the words, but this approach is perilous. If we seek the meaning of "machine" and "think" by examining common usage, we are led to the conclusion that the answer to our question must be sought through a statistical survey, such as a Gallup poll. This is absurd. Instead of attempting such a definition, I shall replace the question with another, closely related one that is expressed in relatively unambiguous words.
\item \textbf{DeepSeek-V3.2 \& Rewrite the following passage (provide only the result):}To address the question "Can machines think?" we must first define the terms "machine" and "think." Relying on common usage, however, is problematic, as it would reduce the inquiry to a mere statistical survey, which is an absurd approach. Therefore, I will not attempt such definitions but will instead replace the original question with a different, more precisely worded one that is closely related.
\item \textbf{DeepSeek-R1 \& Rewrite the following passage (provide only the result):}To address the question "Can machines think?" we must first define the terms "machine" and "think." Using their common meanings, however, is problematic, as it would reduce the inquiry to a matter of statistical opinion, which is an unsatisfactory approach. Therefore, rather than defining these terms, I will replace the original question with a different one that is closely related but phrased more precisely.
\item \textbf{DeepSeek-V3.2 \& Please polish the following passage (provide only the result): }I propose to address the question, "Can machines think?" To do so, one might begin by defining the terms "machine" and "think." Ideally, these definitions would reflect the ordinary usage of the words. However, this approach is problematic. If we seek the meaning of "machine" and "think" solely by observing common usage, we are led to the untenable conclusion that the answer must be determined through something like a statistical survey—a Gallup poll, for instance. This is plainly absurd. Rather than pursuing such definitions, I will instead replace the original question with another, closely related to it, but framed in more precise language.
\item \textbf{DeepSeek-R1 \& Please polish the following passage (provide only the result): }I propose to address the question, "Can machines think?" This inquiry should logically begin by defining the terms "machine" and "think." While one might frame these definitions to reflect the words' ordinary usage, this approach is problematic. If we seek the meaning of "machine" and "think" solely through common usage, we are led to the conclusion that the answer must be found via a statistical survey, like a Gallup poll. This, however, is absurd. Rather than attempting such a definition, I will replace the original question with another that is closely related but framed in more unambiguous terms.
\item \textbf{DeepSeek-V3.2 \& Please rewrite the following passage (provide only the result): }To address the question "Can machines think?" we would first need to define the terms "machine" and "think." One could base these definitions on common usage, but this approach is problematic, as it would reduce the inquiry to a mere statistical survey of popular opinion—an absurd outcome. Rather than pursuing such definitions, I will instead substitute the original question with a different one that is closely related but framed in more precise language.
\item \textbf{DeepSeek-R1 \& Please rewrite the following passage (provide only the result): }I intend to examine the question, "Can machines think?" which ordinarily would start with defining the terms "machine" and "think." While definitions should ideally align with common usage, this method is fraught with risk. If we derive meaning from how words are typically used, we are led to the untenable conclusion that the answer must be found through a statistical survey, such as a Gallup poll. This is clearly absurd. Instead of pursuing such definitions, I will replace the question with a related one that is expressed in more precise language.
\end{itemize}

\end{document}